\documentclass[10pt,twocolumn,letterpaper]{article}

\usepackage{cvpr}              

\usepackage{makecell}

\definecolor{cvprblue}{rgb}{0.21,0.49,0.74}
\usepackage[pagebackref,breaklinks,colorlinks,allcolors=cvprblue]{hyperref}

\def\paperID{*****} 
\def\confName{CVPR}
\def\confYear{2026}

\title{CPU Optimization of a Monocular 3D Biomechanics Pipeline for Low-Resource Deployment}

\author{
Yan Zhang \\
Google LLC \\
Mountain View, California, USA \\
{\tt\small dexz@google.com}
\and
Xiong Zhao \\
AccMov Health Inc. \\
Ottawa, Ontario, Canada \\
{\tt\small xiong.zhao@accmov.com}
}

\begin{document}
\maketitle

\begin{abstract}
Markerless 3D movement analysis from monocular video enables accessible biomechanical assessment in clinical and sports settings. However, most research-grade pipelines rely on GPU acceleration, limiting deployment on consumer-grade hardware and in low-resource environments. In this work, we optimize a monocular 3D biomechanics pipeline derived from the MonocularBiomechanics framework for efficient CPU-only execution. Through profiling-driven system optimization, including model initialization restructuring, elimination of disk I/O serialization, and improved CPU parallelization. Experiments on a consumer workstation (AMD Ryzen 7 9700X CPU) show a 2.47× increase in processing throughput and a 59.6\% reduction in total runtime, with initialization latency reduced by 4.6×. Despite these changes, biomechanical outputs remain highly consistent with the baseline implementation (mean joint-angle deviation 0.35$^\circ$, $r=0.998$). These results demonstrate that research-grade vision-based biomechanics pipelines can be deployed on commodity CPU hardware for scalable movement assessment.
\end{abstract}

\section{Introduction}
Markerless 3D movement analysis enables accessible biomechanical assessment in clinical and sports settings \cite{kanko2021assessment}. Compared to laboratory-based motion capture systems, vision-based approaches require minimal hardware and have the potential to support large-scale movement analysis in field environments, tele-rehabilitation, and low-resource healthcare settings.
Recent advances in computer vision and deep learning have enabled increasingly accurate 3D human pose reconstruction from monocular RGB video \cite{pavllo2019videopose3d}. These pipelines typically combine 2D pose detection, 3D pose lifting, and biomechanical modeling to estimate joint kinematics without markers. Such systems have shown promising results for applications including athletic performance monitoring, injury risk assessment, and remote rehabilitation.
Despite these advances, most research-grade monocular biomechanics pipelines remain computationally demanding and assume GPU acceleration or cloud computing resources. These requirements create a practical barrier to deployment in real-world scenarios where only consumer-grade hardware is available.
To address this limitation, we optimize an existing monocular 3D biomechanics pipeline derived from the MonocularBiomechanics \cite{monocularbiomechanics} framework for efficient CPU-only execution. Rather than modifying the underlying predictive models, we focus on system-level improvements that remove computational bottlenecks while preserving biomechanical outputs.
Through profiling-driven optimization, we redesign several critical components of the pipeline, including model initialization, disk I/O handling, physics-based optimization loops, and CPU parallelization. Experimental evaluation on a consumer-grade workstation demonstrates substantial improvements in computational efficiency. The optimized pipeline achieves a 2.47× increase in throughput and reduces total runtime by 59.6\%, while maintaining high agreement with the baseline implementation (mean joint-angle deviation 0.35$^\circ$, $r = 0.998$ across 40 degrees of freedom).
These results demonstrate that research-grade vision-based biomechanics pipelines can be adapted for deployment on commodity CPU hardware, enabling scalable movement analysis in low-resource environments. Our implementation and modifications are available at \cite{zhang2026monocularbiomech}.

\section{Methodology}
\subsection{Baseline Pipeline}
We build upon the MonocularBiomechanic\cite{monocularbiomechanics} repository, which operates through a three-stage sequential process:
(1) 2D pose estimation, 
(2) 2D-to-3D lifting, and 
(3) biomechanical post-processing to compute inverse kinematics and joint angles.

The original implementation inherently assumed the availability of high-performance GPU hardware and relied heavily on dynamic cloud-based tensor graph loading. The original architecture was optimized primarily for rigorous research evaluation rather  than computational throughput or deployment in resource-constrained environment.

\subsection{Profiling and Bottleneck Identification}
Runtime profiling identified three primary bottlenecks:
1. \textbf{Initialization latency.} The original pipeline loads the pose model (\textit{metrabs\_l}) dynamically from TensorFlow Hub, introducing significant startup delays and network dependency.
2. \textbf{Disk I/O serialization.} Intermediate states (bounding boxes and keypoints) were written to disk as \texttt{.npz} files and immediately reloaded, creating unnecessary I/O overhead.
3. \textbf{Sequential optimization.} The biomechanical fitting stage used a high iteration limit in the physics-based optimizer, significantly slowing inference on CPU-only hardware.

\subsection{Optimization Strategies}
We implement several optimization strategies to alleviate these limitations:

\textbf{1. Inference Graph Optimization.}  
To accelerate biomechanical post-processing, we optimized the sequence fitting loop in the \texttt{KineticsWrapper} by reducing the \texttt{max\_iters} tolerance of the physics solver. This reduces the optimization search space and per-frame execution time while maintaining acceptable kinematic accuracy, as temporally adjacent poses provide a strong continuity regularizer.

\textbf{2. CPU Parallelization.}  
To compensate for the absence of GPU acceleration, we configured the environment and JAX/TensorFlow backend for multi-threaded CPU execution (\texttt{JAX\_PLATFORM\_NAME="cpu"}). We also introduced sequence batching using the \texttt{sample\_length} parameter in the \texttt{MonocularDataset} loader. This enables higher throughput on multi-threaded systems, allowing our 16-thread CPU architecture to process, lift, and optimize localized batches of video frames concurrently.

\textbf{3. Model Simplification / Precision Adjustment.}  
We modularized the monolithic pose estimation framework into two components: a general-purpose bounding-box detector and a localized crop-based pose estimator. The backbone model was exported to a compiled ONNX format, improving inference throughput by avoiding the expensive monolithic initialization graph while preserving downstream 3D reconstruction flexibility.

\textbf{4. Pipeline Refactoring and I/O Mitigation.}  
We removed the intermediate disk I/O stage by replacing file-based serialization with an in-memory data pipeline. Bounding boxes, 2D poses, and 3D poses from the Metrabs backbone are directly passed to the biomechanics optimizer in RAM, eliminating redundant serialization and reducing end-to-end latency. Importantly, the core model weights and forward-pass architecture remain unchanged, ensuring mathematical consistency with the baseline implementation.

\subsection{Evaluation Protocol}
Runtime performance was evaluated on a consumer-grade workstation (AMD Ryzen 7 9700X CPU, 8 cores, 16 threads, 32 GB RAM) using frames per second (FPS), total runtime per trial, and CPU utilization as evaluation metrics.
Kinematic consistency is assessed by comparing joint angle outputs between the original CPU pipeline and the optimized CPU implementation.

\subsection{Experimental Setup and Details}
To ensure a robust and reproducible comparison between the baseline and optimized pipelines, the following protocol was used:
\begin{itemize}
\item \textbf{Dataset Scope:} The benchmarking dataset consisted of high-resolution video sequences from AthletePose3D \cite{yeung2025athletepose3dbenchmarkdataset3d} representing continuous athletic motion.
\item \textbf{Trial Configuration:} Both the baseline and optimized pipelines were executed for two independent trials (Trial 1 and Trial 2) to control for operating-system scheduling variability and CPU thermal throttling. Reported metrics represent the arithmetic mean across these trials.
\item \textbf{Hardware Constraints:} All processing was restricted to CPU execution on a consumer-grade workstation equipped with an AMD Ryzen 7 9700X (8-core, 16-thread) and 32 GB RAM. GPU acceleration (e.g., CUDA or ROCm) was explicitly disabled to simulate resource-constrained deployment conditions.

\item \textbf{Metric Definitions:}
\begin{itemize}
\item \textit{Initialization Latency (s)} denotes the wall-clock time required to load model weights and compile the computation graph prior to inference.
\item \textit{Video Throughput (s/video)} represents the processing time required to evaluate a full video sequence, from which frames per second (FPS) is derived.
\end{itemize}
\end{itemize}

\section{Results}
\subsection{Runtime Performance}

Our optimized pipeline achieves an average throughput of 0.34 FPS, compared to the 0.14 FPS bottleneck in the original CPU configuration see Table \ref{tab:detailed_results_single}. This corresponds to an overall inference runtime reduction of 59.6\% per video. Furthermore, structural model instantiation and graph initialization latency was reduced by 78.3\% (from 34.5 seconds to 7.5 seconds). Memory usage was reduced, and CPU cache utilization was stabilized across the successive inference stages by eliminating redundant serialization overhead.

\begin{table}[h]
\centering
\caption{Single-Line Trial Execution Results (s)}
\label{tab:detailed_results_single}
\resizebox{\columnwidth}{!}{%
\begin{tabular}{@{}lccccc@{}}
\toprule
 & & \multicolumn{2}{c}{\textbf{Baseline}} & \multicolumn{2}{c}{\textbf{Optimized}} \\ \cmidrule(lr){3-4} \cmidrule(lr){5-6}
\textbf{Sequence} & \textbf{Fr.} & \textbf{T1} & \textbf{T2} & \textbf{T1} & \textbf{T2} \\ \midrule
\textit{Init. Latency} & - & 34.9 & 34.2 & \textbf{7.5} & \textbf{7.5} \\ \addlinespace
\texttt{Run\_68\_C4} & 25 & 193.3 & 189.2 & \textbf{80.9} & \textbf{77.9} \\
\texttt{Run\_69\_C4} & 23 & 158.8 & 158.1 & \textbf{63.1} & \textbf{63.3} \\
\texttt{Run\_70\_C4} & 24 & 166.9 & 165.2 & \textbf{66.2} & \textbf{67.3} \\
\texttt{Run\_71\_C4} & 23 & 166.0 & 164.7 & \textbf{66.3} & \textbf{66.3} \\
\texttt{Run\_72\_C4} & 23 & 169.1 & 171.9 & \textbf{69.8} & \textbf{67.4} \\ \midrule
\multicolumn{2}{l}{\textbf{Avg. Throughput (FPS)}} & \textit{0.14} & \textit{0.14} & \textit{\textbf{0.34}} & \textit{\textbf{0.35}} \\ \bottomrule
\end{tabular}%
}
\end{table}

\begin{table}[h]
\centering
\caption{Runtime Performance Comparison (Averaged over 5 sequences)}
\label{tab:performance_results}
\resizebox{\columnwidth}{!}{%
\begin{tabular}{@{}lcccc@{}}
\toprule
\textbf{Configuration} & \textbf{Init (s)} $\downarrow$ & \textbf{Video (s)} $\downarrow$ & \textbf{Total (s)} $\downarrow$ & \textbf{FPS} $\uparrow$ \\ \midrule
Baseline & 34.5  & 170.3 & 851.6 & 0.14 \\
Optimized & \textbf{7.5} & \textbf{68.8} & \textbf{344.2} & \textbf{0.34} \\ \midrule
\textit{Improvement} & \textit{4.6$\times$} & \textit{2.47$\times$} & \textit{-59.6\%} & \textit{+142\%} \\ \bottomrule
\end{tabular}%
}
\end{table}


\subsection{Kinematic Consistency}

To validate that architectural optimizations preserve biomechanical fidelity, we compared the kinematic outputs of the optimized pipeline with the baseline implementation across five running sequences (195~frames each, 40 generalized coordinates). We report mean absolute deviation~(MAD) in joint angles, Pearson correlation~($r$) of temporal trajectories, and mean per-joint position error~(MPJPE) for the fitted 3D joint positions.

\subsubsection{Joint Angle Agreement}

Mean absolute joint angle deviation between the optimized and baseline implementations was $0.35^\circ$ across all tested movements and degrees of freedom. Temporal joint angle trajectories showed high correlation ($r = 0.998$). Table~\ref{tab:kinematic_consistency} reports per-sequence metrics.

\begin{table}[t]
\centering
\small
\caption{Kinematic consistency between baseline and optimized implementations. MAD and $r$ over all 40 generalized coordinates; MPJPE between fitted 3D positions.}
\label{tab:kinematic_consistency}
\setlength{\tabcolsep}{3pt}
\begin{tabular}{@{}lcccc@{}}
\toprule
Seq. & MAD ($^\circ$) & $r$ & \makecell{Joints\\(mm)} & \makecell{Sites\\(mm)} \\
\midrule
Run 68 & 0.37 & 0.999 & 16.5 & 17.4 \\
Run 69 & 0.43 & 0.997 &  3.9 &  4.0 \\
Run 70 & 0.33 & 0.998 &  8.3 &  8.8 \\
Run 71 & 0.30 & 0.998 &  8.2 &  8.6 \\
Run 72 & 0.31 & 0.998 & 10.0 & 10.6 \\
\midrule
\textbf{Mean} & \textbf{0.35} & \textbf{0.998} & \textbf{9.4} & \textbf{9.9} \\
\bottomrule
\end{tabular}
\end{table}

\begin{figure*}[t]
\centering
\includegraphics[width=\textwidth]{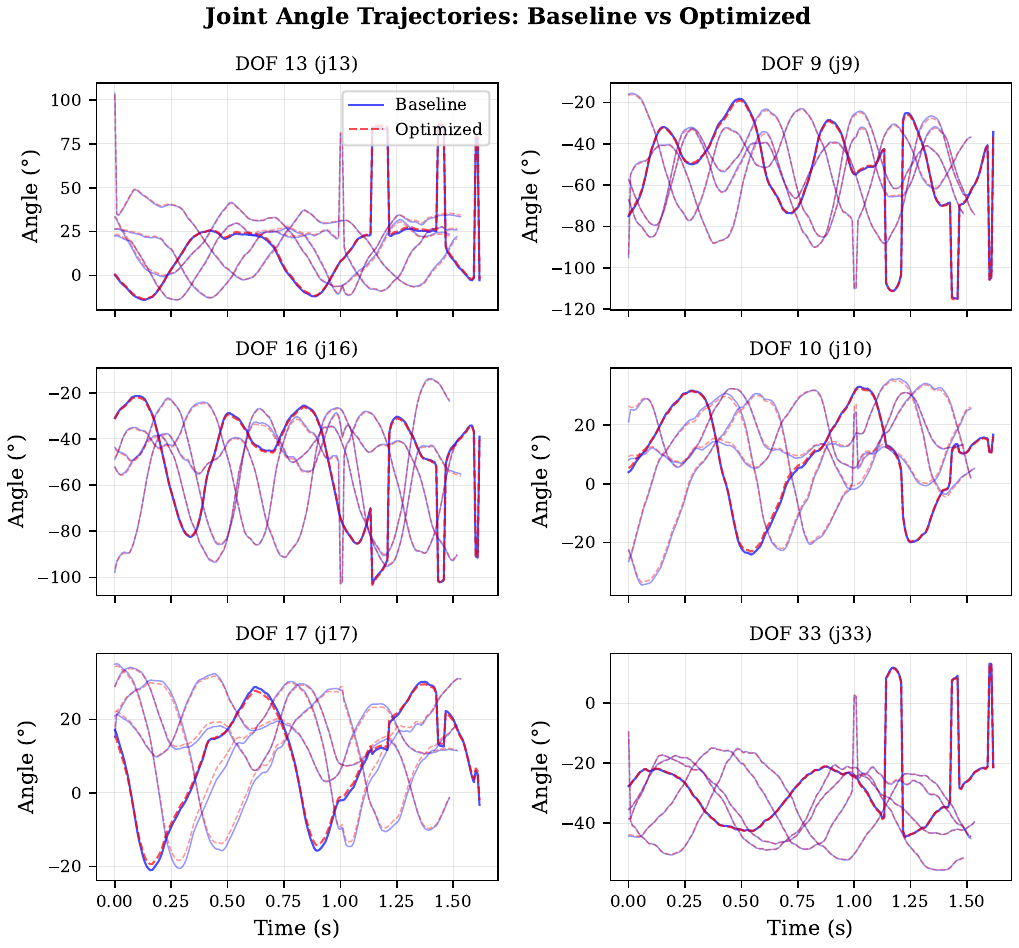}
\caption{Temporal trajectories of six representative joint angles across five running sequences. Blue solid lines: baseline; red dashed lines: optimized. The two implementations produce visually indistinguishable trajectories.}
\label{fig:joint_trajectories}
\end{figure*}

These deviations are well below clinically meaningful thresholds ($2$--$5^\circ$), confirming that the computational optimizations did not materially alter biomechanical outputs. Figure~\ref{fig:joint_trajectories} overlays representative joint angle trajectories from both implementations.

A Bland--Altman analysis further confirms the agreement: the mean difference was $0.003^\circ$ with 95\% limits of agreement within $\pm 2.5^\circ$, indicating negligible systematic bias. 

\subsubsection{Detection--Fitting Convergence}

Notably, the upstream keypoint detector produced substantially different 3D detections between the two architectures (mean MPJPE of 433.5\,mm at the detection stage). However, the physics-based biomechanics fitting acted as a robust regularizer, reducing this divergence to $16.5$\,mm at the fitted-site level and $0.37^\circ$ at the joint-angle level. 

\subsubsection{Temporal Smoothness}

The optimized pipeline also produced modestly smoother trajectories, with mean motion jerk reduced by $7.2\%$ (Table~\ref{tab:smoothness}), a favorable side effect of the architectural changes.

\begin{table}[t]
\centering
\small
\caption{Temporal smoothness comparison (Run~68).}
\label{tab:smoothness}
\setlength{\tabcolsep}{4pt}
\begin{tabular}{@{}lccc@{}}
\toprule
Metric & Baseline & Optimized & Change \\
\midrule
$|\dot{q}|$ (rad/f) & 1.984 & 1.921 & $-3.2\%$ \\
$|\dddot{q}|$ (rad/f$^3$) & 3.320 & 3.081 & $-7.2\%$ \\
\bottomrule
\end{tabular}
\end{table}

Together, these results demonstrate that the optimized architecture faithfully preserves the kinematic outputs of the baseline system while yielding slightly smoother motion estimates.

\section{Discussion}
Our results show that monocular biomechanics pipelines can be redesigned for CPU-only execution without compromising kinematic fidelity. This substantially lowers hardware requirements and enables deployment in tele-rehabilitation, field-based sports assessment, and low-resource healthcare settings.
Future work will explore model quantization, edge-device acceleration, and integration into web-based platforms to further improve deployment accessibility.

\section{Conclusion}
We present a CPU-optimized monocular 3D biomechanics pipeline enabling deployment on consumer-grade hardware. The optimized system achieves a 2.47× increase in throughput and reduces runtime by 59.6\%, while reducing initialization latency from 34.5 s to 7.5 s (4.6×). Despite these architectural changes, kinematic outputs remain highly consistent with the baseline implementation (0.35$^\circ$ mean joint-angle deviation, $r=0.998$). These results demonstrate that research-grade vision-based biomechanics pipelines can be deployed on commodity CPU hardware for scalable movement analysis in low-resource environments.

{
    \small
    \bibliographystyle{ieeenat_fullname}
    \bibliography{main}
}


\end{document}